# HCRS: A hybrid clothes recommender system based on user ratings and product features

*Xiaosong Hu, Wen Zhu, Qing Li*
Sch. of Economic Inf. Eng.
Southwestern Univ. of Finance & Econ.
Chengdu, China

*Abstract*-Nowadays, online clothes-selling business has become popular and extremely attractive because of its convenience and cheap-and-fine price. Good examples of these successful Web sites include Yintai.com, Vancl.com and Shop.vipshop.com which provide thousands of clothes for online shoppers. The challenge for online shoppers lies on how to find a good product from lots of options. In this article, we propose a collaborative clothes recommender for easy shopping. One of the unique features of this system is the ability to recommend clothes in terms of both user ratings and clothing attributes. Experiments in our simulation environment show that the proposed recommender can better satisfy the needs of users.

*Keywords-Collaborative filtering; Clothes recommender system; Probabilistic model; Information filtering*

## I. INTRODUCTION

The concept of a recommender system was first put forward in the mid–1990s [1]. The basic function of a recommender system is to suggest interesting products to end users in terms of users' behavior. With the fast growth of e-commerce, recommenders have been playing a vital role in e-commerce and have attracted great attention from chief-officers in the big e-business companies including Amazon.com, Tmail.com. In factuality, many companies find out that recommenders can not only recommend goods which fit customers, but also take enormous effect in converting browsers into buyers, increasing cross-sell and building loyalty [2]. The adoption of recommender system in business is widespread.

In general, the recommender can be categorized into three categories, i.e., the content-based recommender, the collaborative recommender, and the hybrid recommender.

- The content-based recommender recommends products that are most similar to what the user like most in terms of the inner attributes of the products. It derived from information retrieval with a specific focus on long-term information filtering. A good example is the PicSOM which recommends similar images based on the color, texture, and shape of images [3].
- For the collaborative recommender system [5], it usually recommends products considering the opinions of others. It is a sort of word-of-mouth advertisement. Specifically, it first finds friends of a target user who share similar interest on the same products based on the historical information. And then, it predicts the preference of the target user on a certain product based on the opinions of his/her friends [4]. It can be further categorized into memory-based recommender and model-based recommender [6].
- The hybrid system takes both forms of the content-based system and the collaborative system [6]. It recommends products based on both the product attributes and user ratings. A good example of such hybrid recommender is the music recommender system developed by Li et al. [7], which suggests music in terms of user ratings and audio features.

In this article, we propose a hybrid recommender system for easy clothes shopping. One of the unique features of this system is the ability to recommend clothes in terms of both user ratings and clothing attributes. Experiments in our simulation environment show that the proposed recommender can better satisfy the needs of users.

## II. SYSTEM DESIGN

Figure 1 is the outline of our proposed hybrid clothes recommendation system, dubbed HCRS, which takes the utilization of both user ratings and product features. In particular, it first applies the human detection techniques to detect the clothes area in an image. Second, it analyzes the clothes area and calculates the percentage of each color in this area. Third, it extends the item-rating matrix with group-rating matrix to accommodate product features for recommendation. Fourth, similar products of the target product to be recommended are selected based on such extended rating matrix. At last, the recommended products are determined by the ratings earned by the similar products.

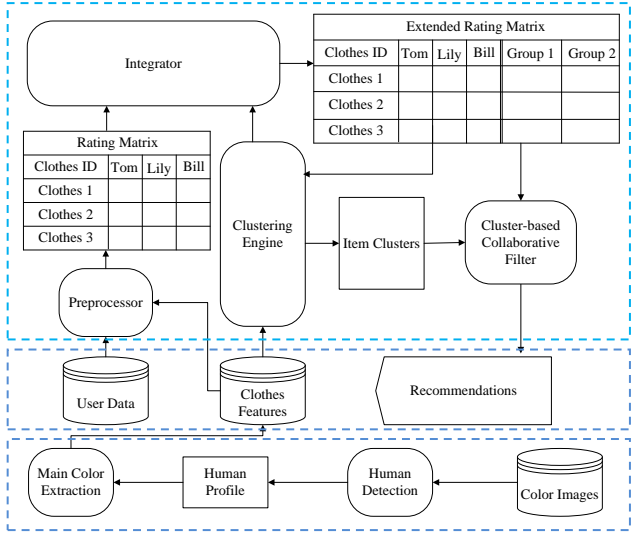

Figure 1. Outline of the HCRS system.

*Algorithm 1: Adjusted K-means Clustering*
*Input : the number of clusters k and item attribute features.*
*Output : a set of k clusters that minimizes of the squared-error criterion, and the probability of each item belonging to each cluster center.*

*(1) Choose k objects as the initial cluster centers;*
*(2) Repeat (a) and (b) until small change;*
  *(a) (Re)assign each object to the cluster to which the object is the most similar, based on the mean value of the objects in the cluster;*
  *(b) Update the cluster means, i.e., calculate the mean value of the objects for each cluster;*
*(3) Compute the possibility between each object and each cluster center.*

Figure 2. Outline of the HCRS system.

| Clothes ID | Tom | Lily | Jack | Bill | Group 1 | Group2 |
|---|---|---|---|---|---|---|
| Clothes 1 | 3 | | | 2 | 0.357 | 0.189 |
| Clothes 2 | | 1 | | | 0.222 | 0.45 |
| Clothes 3 | 3 | | 3 | | 0.12 | 0.789 |
| Clothes 4 | | 2 | | | 0.18 | 0.43 |
| Clothes 5 | 2 | | | | 0.115 | 0.253 |
| Clothes 6 | | | 3 | 1 | 0.367 | 0.322 |

Figure 3. Extend item-rating matrix.

## III. RATING MATRIX CONSTRUCTION

### A. Group-Rating Matrix Construction

User ratings generally are not comprehensive enough. To handle this challenge, we utilize the product features as a complement to capture the inner connections of products. More specifically, we first group clothes into clusters based on the product features, each clothes belongs one or more product groups. Such memberships are utilized as "group-ratings" to construct an extended rating matrix. Essentially, these "group-ratings" provide abstract summarization of the primitive low-level features, revealing the high-level semantic information.

The traditional K-means clustering [8] is a hard clustering technique, in which each item only can be assigned into one cluster. In this study, we extend it with fuzzy set theory to support soft clustering, in which each item belongs to one or more groups, for our recommendation purpose. The details of this adjusted K-means algorithm is shown in Figure 2. In particular, the probability of one product assigned to a certain clique k is defined as:

$$Pro(j,k) = 1 - \frac{CS(j,k)}{MaxCS(i,k)} \quad (1)$$

where $Pro(j,k)$ is the possibility of object j belonging to cluster k; $CS(j,k)$ is the counter-similarity between the object j and the cluster k, which is calculated based on the Euclidean distance; and $MaxCS(i,k)$ is the maximum counter-similarity between an object and cluster k.

### B. Extend Item-Rating Matrix

With the group-ratings, we extend the traditional item-rating matrix for recommending products. The group-ratings are attached to the item-ratings as "pseudo-ratings" as shown in Figure 3.

### C. Collaborative Recommendations

To determine whether a product should be recommended or not, HCRS takes collaborative filtering techniques to predict the probability of recommendation. In particular, the prediction for recommending a product is calculated by performing a weighted average of deviations from the neighbors' mean as

$$P_{k,i} = \overline{R_k} + \frac{\sum_{u=1}^{n}(R_{u,i}-\overline{R_u}) \times sim(k,u)}{\sum_{u=1}^{n}|sim(k,u)|}, \quad (2)$$

where $P_{k,i}$ represents the prediction for user *i* of item *k*, *n* denotes the top *N* nearest neighbors of item *k*, $\overline{R_k}$ and $\overline{R_u}$ represent the average ratings of item *k* and *u*, respectively.

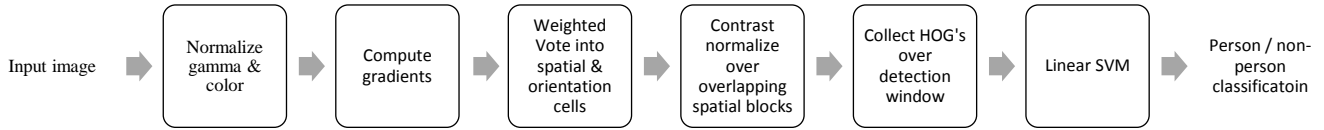

Figure 4. An overview of the human detection chain

## IV. FEATURE EXTRACTION

A user's preference of clothes color is a key factor for his or her shopping decision-making [9]. In this section, we describe the approach to extract physical features from clothes images to provide additional information for clothes recommendation. Specifically, we first use histograms of oriented gradients [10] to detect the profile of human being in a cloth image, and then analyze dominant colors from the extracted area.

### A. Human Detection

In this section, we adopt the method proposed by Navneet Dalal and Bill Triggs [10] to detect the profile of fashion models in clothes image (Figure 4). Specifically,

1. Gamma/Color Normalization: First, we normalize an image for later feature detection and extraction. However, gamma coding was put forward to compensate for the properties of human vision, hence to maximize the use of the bits or bandwidth relative to how humans perceive light and color. And it is widely used in study of photography. According to the work of Navneet Dalal and Bill Triggs [10], RGB color space with no gamma correction displays superior performance comparing to the normalizations with grayscale, RGB and LAB. Therefor, we use RGB color method in this work.
2. Gradient Computation: An image gradient is a directional change in the intensity or color in the image which is widely used to extract information from images. In this part, we use [-1, 0, 1] gradient filter with no smoothing to calculate gradient for each color channel in the color image, and take the one with the largest norm as the pixel's gradient vector to enhance the sensitivity.
3. Spatial / Orientation Binning: After step 2 we found that gradient strengths vary over a wide range owing to local variations, so effective local contrast normalization turns out to be essential for good performance. In the experiment, we take $8\times 8$ pixel as a cell. And for each cell, we apply the method of linear gradient voting into 9 orientation bins in $1°-180°$ to get cells.
4. Normalization and descriptor blocks: To obtain useful information from larger area, we group four $8\times 8$ pixel cells into $16\times 16$ pixel blocks and get blocks with block spacing stride of 8 pixels (hence 4-fold coverage of each cell). And then we use L2-Hys to normalize blocks.
5. Detector Window and Context: we use $64\times 128$ pixel to be a detector window to represent a part of an image.
6. SVM is one of the most popular algorithms for pattern recognition. As we mention before, in order to detect human being, we link SVM with the human-image and background-image to training a model for human detection.

### B. Main Color Extraction

According to biology, human being cannot distinguish more than 150 colors. HSV which is short for hue, saturation, and value is formally described by Alvy Ray Smith [11]. It can be converted from RGB (red, green and blue) to represent a color with an array consisted of 3 numbers. And it is more intuitive and perceptually relevant than the Cartesian representation. Therefore, we can use different combination to represent different colors that can be distinguished by human being.

However, in the theory of HSV, hue has a greater influence on color detecting than saturation and value. So we use Equation 3 and equation in Figure 5 to generate the number that can present a color.

$$L = 9H + 3S + V \tag{3}$$

where L stands for the number representing a set of color which can not be distinguished by human. And H, S and V stand for the result of the function in Figure 5.

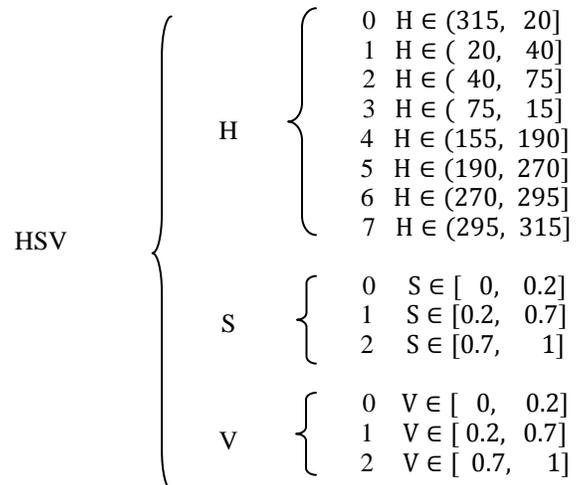

Figure 5. Way of caculating the color

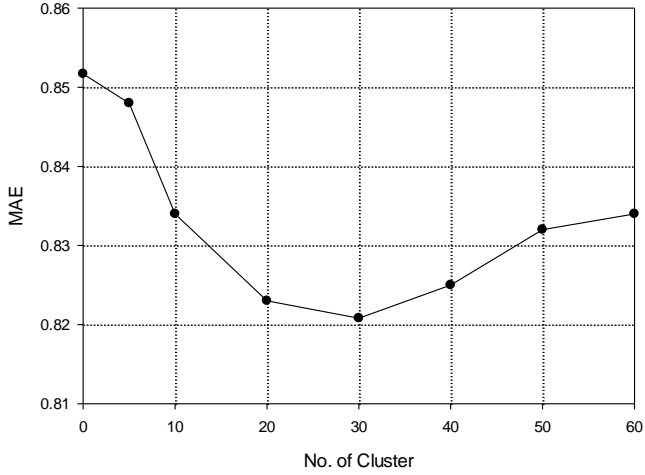

Figure 6. Clustering

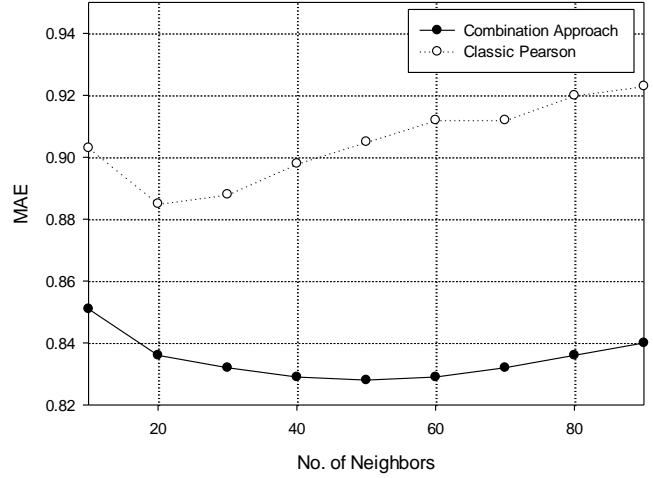

Figure 7. Comparison of our approach and classic Pearson method

## C. Vector for Clothes

In this section, we calculate L for each of the pixels in the picture. And then, we use Equation 4 to calculate the percentage for each color (represented by L in Equation 3) in each picture.

$$\text{Per}(\text{color}_n) = \frac{\text{count}(\text{pixels}, L = L_n)}{\text{count}(\text{pixels}, \text{in the picture})} \quad (4)$$

$\text{Per}(\text{color}_n)$ stands for the percentage of the kind of color in the image where $\text{count}(\text{pixels}, L = L_n)$ stands for the number of pixels belonging to this kind of color and $\text{count}(\text{pixels}, \text{in the picture})$ stands for the total amount of the pixels in the image.

## V. EXPERIMENT EVALUATIONS

To gauge the performance of our approach, we carried out a series of experiments.

### A. Data Set & Evaluation Metric

We performed experiments on the data set generated by 163 users who usually purchase their new clothes on e-commercial Web sites. Every user is supported to give a mark on 50 clothes of different styles. However, after wiping out the blank record, we achieve a total of 1783 rating record. And then we divide the data set into a training set and a test data set. Twenty percent of users are randomly selected to be the test users and the rest servers as the training set.

MAE [4] is used as the metric in our experiments, which is calculated by summing these absolute errors of corresponding rating–prediction pairs and then computing the average. A lower MAE indicates greater accuracy.

$$\text{MAE} = \frac{\sum_{k=1}^{N} |P_{i,k} - R_{i,k}|}{N} \quad (5)$$

where $P_{i,k}$ is the system's rating (prediction) of item k for user i, and $R_{i,k}$ is the rating of the user i for item k in the test data. N is the number of rating-prediction pairs between the test data and the prediction result.

### B. Behaviors of The Proposed Recommender

*1)* Number of clusters: We implement group-rating method described as in part III and found that the number of clusters to a massive extent as show in Figure 6. It can be observed that the number of clusters affects the quality of prediction. And when the number of clusters is about 30, MAE reach the optimal value.

*2)* Methods for computing user-user similarity: As we have observed, the value scale of the group-rating matrix and user-rating is distinct. Hence, we should modify the value of the same scale or separately compute the user-user similarity. In our experiments, we enlarge the value scale of the group-rating matrix from [0 1] to [0 5]. And then we use the Pearson correlation-based algorithm to calculate the user-user similarity from the user-rating matrix and adjusted cosine algorithm for the similarity from the group-rating matrix. Finally, the total user-user similarity is then calculated as the linear combination of the above two. And Figure 7 shows that our linear combination method displays better performance than then classic algorithm.

*3)* Neighborhood size: The size of the neighborhood has a significant effect on the prediction quality [12]. In our experiment, we vary the number of neighbors and compute the MAE. It can be observed from Figure 7 that the size of the neighbors affect the quality of the prediction. When the number of neighbors is changed from 40 to 60 in our approach, the optimal MAE value is obtained.

## VI. CONCLUTION

In this article, we propose a hybrid recommender system for easy clothes shopping. One of the unique features of this system is the ability to recommend clothes in terms of both user ratings and clothing attributes. To simplify your challenge, we only extract the color information of clothes to enrich the experience learning of recommender system. Experiments in our simulation environment show that the proposed recommender can better satisfy the needs of users. However, it would be interesting to explore more features of clothes including clothe texture or even fashion style.

## VII. ACKNOWLEDGMENT

This work has been supported by the National Natural Science Foundation of China (NSFC Grant No. 60803106, 61170133, 60903201, 91218301), the Fok Ying-Tong Education Foundation, China (Grant No. 121068), the Program for Key Innovation Research Team in Sichuan Province (2011JTD0028).